\documentclass[letterpaper, 10 pt, conference]{ieeeconf}  

\IEEEoverridecommandlockouts                              

\usepackage{scalerel}
\newcommand\llbracket{\big[\!\!\big[}
\newcommand\rrbracket{\big]\!\!\big]}

\usepackage{graphicx}
\usepackage{cite}
\usepackage{amsmath,amsfonts}
\usepackage[hidelinks]{hyperref}
\hypersetup
{
colorlinks = true,
linkcolor = red,
anchorcolor = black,
citecolor = green,
urlcolor = magenta
}

\title{\LARGE \bf
Monocular Visual Place Recognition in LiDAR Maps via Cross-Modal State Space Model and Multi-View Matching
}

\author{Gongxin Yao, Xinyang Li, Luowei Fu and Yu Pan$^{*}$
\thanks{Gongxin Yao, Xinyang Li, Luowei Fu and Yu Pan are with the Institute of Cyber-Systems and Control, Zhejiang University, Hangzhou, 310027, China. (Yu Pan* is the corresponding author, email: ypan@zju.edu.cn)}%
}

\begin{document}

\maketitle
\thispagestyle{empty}
\pagestyle{empty}

\begin{abstract}
Achieving monocular camera localization within pre-built LiDAR maps can bypass the simultaneous mapping process of visual SLAM systems, potentially reducing the computational overhead of autonomous localization. To this end, one of the key challenges is cross-modal place recognition, which involves retrieving 3D scenes (point clouds) from a LiDAR map according to online RGB images. In this paper, we introduce an efficient framework to learn descriptors for both RGB images and point clouds. It takes visual state space model (VMamba) as the backbone and employs a pixel-view-scene joint training strategy for cross-modal contrastive learning. To address the field-of-view differences, independent descriptors are generated from multiple evenly distributed viewpoints for point clouds. A visible 3D points overlap strategy is then designed to quantify the similarity between point cloud views and RGB images for multi-view supervision. Additionally, when generating descriptors from pixel-level features using NetVLAD, we compensate for the loss of geometric information, and introduce an efficient scheme for multi-view generation. Experimental results on the KITTI and KITTI-360 datasets demonstrate the effectiveness and generalization of our method. The code will be released upon acceptance.

\end{abstract}

\section{INTRODUCTION}

Autonomous localization is essential for mobile robots, with Simultaneous Localization and Mapping (SLAM) system\cite{placed2023survey} being the most classic solution. However, the mapping process in camera-based visual SLAM \cite{fuentes2015visual} is quite complex. After matching keypoints and reconstructing 3D information through triangulation \cite{campos2021orb}, it typically relies on graph optimization \cite{grisetti2010tutorial, kummerle2011g} to continuously refine the trajectory and 3D map. In addition to the significant computational costs, the completeness and distance accuracy of the 3D map are often inferior to those obtained from LiDAR measurements. LiDAR-based SLAM \cite{zhang2014loam, lee2024lidar} still requires extensive mapping steps and also increases the hardware cost, which is prohibitive for some micro robots. Therefore, some researches \cite{caselitz2016monocular, yu2020monocular} combine their respective advantages by using a low-cost camera for mapping-free localization within pre-built and high-precision LiDAR maps. 

\begin{figure}[t]
    \centering
    \includegraphics[width=1\linewidth]{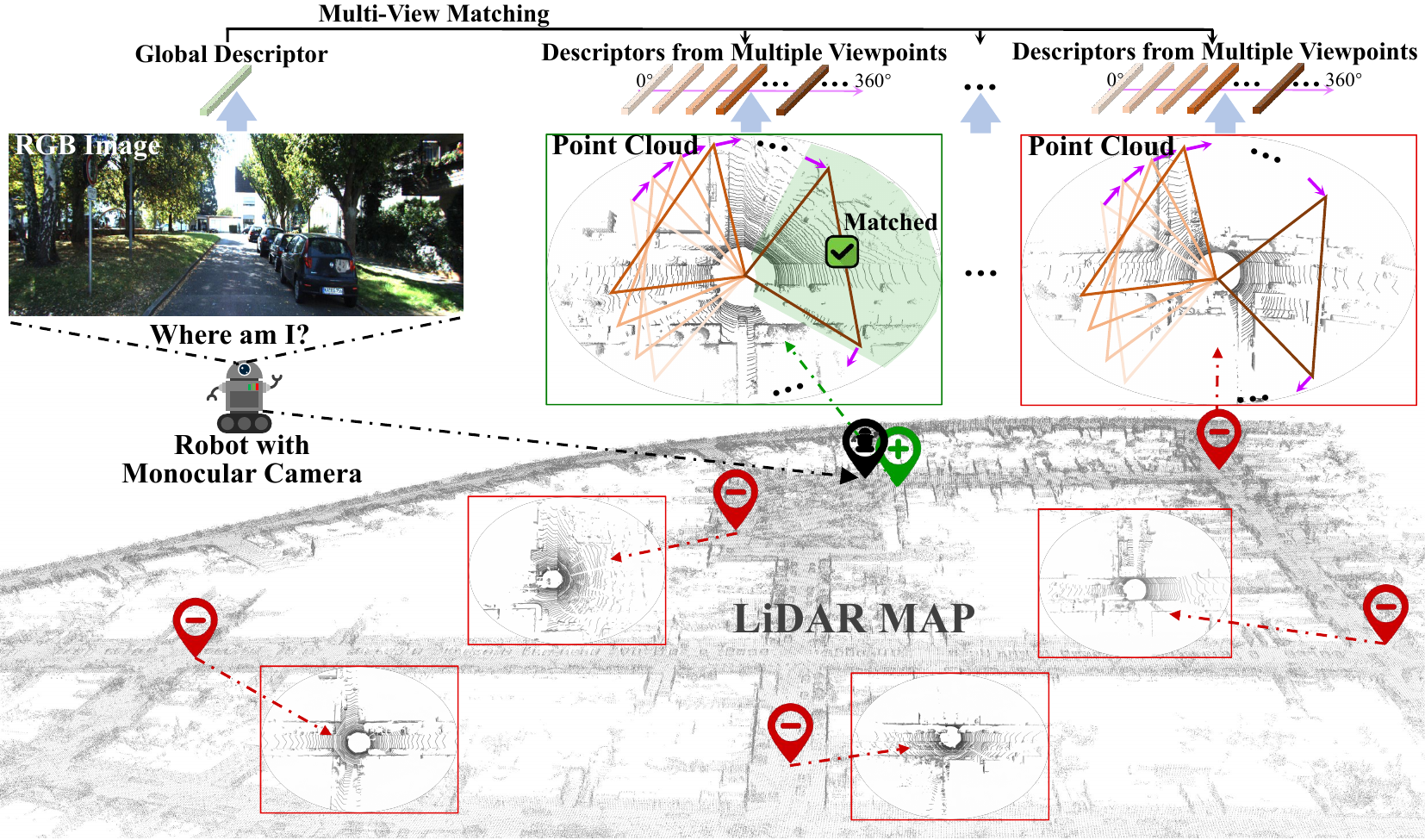}
    \vspace{-5mm}
    \caption{Illustration of the cross-modal place recognition, which enables a robot with monocular camera to localize within a pre-built LiDAR map. We generate a global descriptor for online RGB image while independent descriptors from multiple viewpoints for point clouds. It bridges the modality gaps and reduces the FOV differences, improving retrieval accuracy.}
    \vspace{-4mm}
    \label{fig1}
\end{figure}

Place recognition is one of the key back-end components in SLAM systems, identifying the revisited place for loop closure optimization \cite{placed2023survey}. This process typically involves generating and retrieving descriptors \cite{zhang2021visual, luo20243d} within a single modality. In contrast, image-to-point-cloud place recognition enables direct camera localization within LiDAR maps, but is challenged by the modality gaps. To address it, recent works \cite{lee20232, zheng2023i2p} reconstruct depth information from RGB images before generating descriptors, and others \cite{cattaneo2020global, xie2024modalink, li2024vxp, feng20192d3d, yao2023cfi2p} directly learn a unified embedding space for both modalities. In general, these methods adopt two main-stream deep learning architectures, convolutional neural networks (CNNs) \cite{li2021survey} or vision transformers (ViTs) \cite{han2022survey}, for feature extraction. While CNNs are limited by the local receptive field of convolution kernels, ViTs can model long-range relationships but introduce high computational overhead due to the quadratic complexity of self-attention \cite{vaswani2017attention}. A model that balances both efficiency and effectiveness has long been needed.

Field-of-view (FOV) differences further degrade the quality of global descriptors in cross-modal place recognition. Since 360° point clouds often contain irrelevant scene information outside the FOV of RGB images, retrieving scenes with global descriptors \cite{cattaneo2020global, li2024vxp} suffers from low accuracy. To address it, some works \cite{zheng2023i2p, xie2024modalink} use camera pose priors to crop the point clouds. However, such cropping is not feasible in practice since the camera poses relative to LiDAR maps are unknown. Besides, NetVLAD \cite{arandjelovic2016netvlad} learns multiple semantic cluster centers and aggregates local feature residuals into a global descriptor \cite{ZHANG2021107952, fu2024coarse}. Empirically, we observed that NetVLAD often retrieves incorrect scenes that are semantically similar to the target but differ in geometry. This occurs because pixel-wise residual summation disrupts the distribution of image content (see Section \ref{sec:netvlad}), leading to a loss of geometric information. 


In this paper, we present an efficient framework based on visual state space model (VMamba) \cite{liu2024vmamba} to jointly learn descriptors for RGB images and point clouds. VMamba is a recursive architecture with linear complexity. It controls the information interaction between sequential visual tokens via a parallel selective scanning mechanism \cite{gu2023mamba}, thereby capturing long-range relationships. For a compact and unified input representation, point clouds are converted to 360° range images. A single inference can produce independent descriptors for multiple evenly distributed views in the range images, each with an FOV comparable to that of the RGB images. Intuitively, fine-grained multi-view matching helps mitigate the impact of irrelevant scenes, as shown in  \mbox{Fig. \ref{fig1}}. Afterwards, it is crucial to quantify the similarity between the point cloud views and RGB images for multi-view supervision. Since the numerical integration of FOV overlap is complex, we propose a 3D visible point-based method that is both computationally efficient and closely tied to the visual content. First, surface points corresponding to RGB images are truncated from the 3D map, and then the visible ratio of these points from different viewpoints is calculated via pose transformation. Beyond multi-view contrastive learning, we also propose a joint supervision across pixels, views, and scenes to enhance cross-modal representations. Finally, when generating descriptors from pixel-level features using NetVLAD, we compensate for the geometric information loss by fusing a convolutional branch, and introduce an efficient multi-view generation scheme for the 360° range images. In conclusion, our main contributions are:
\begin{itemize}
    \item We propose a VMamba-based framework to generate global descriptors for RGB images and multi-view descriptors for point clouds, with a pixel-view-scene joint training strategy for cross-modal contrastive learning.
    \item We propose a 3D visible point-based strategy to efficiently quantify the similarity between RGB images and point cloud views for multi-view supervision.
    \item We expose the limitations of vanilla NetVLAD, enhance its performance for place recognition through geometric compensation, and introduce an efficient scheme for generating multi-view descriptors.
\end{itemize}

Extensive experiments on KITTI \cite{geiger2013vision} and KITTI-360 \cite{liao2022kitti} datasets show that our method outperforms the state-of-the-art methods in cross-modal place recognition. Our method can simultaneously generate a global descriptor for an RGB image and multi-view descriptors for a 360° range image in just 25.31ms ($\sim$39Hz) on a 3090 GPU (see Section \ref{sec:ablation}).


\section{Related works}

\subsection{Localization and Place Recognition}
Traditional two-step localization pipeline \cite{sattler2016large, torii2019large} first use place recognition for coarse localization and then apply data registration \cite{ma2021image, huang2021comprehensive} for finer pose estimation. We recommend these papers \cite{zhang2021visual, luo20243d} for an overview of place recognition in single-modal data. Here, we mainly focus on reviewing image-to-point cloud place recognition methods, which can be classified into two main categories. One line tends to mitigate the inherent modality gaps via some explicit operations. For example, LC$^2$ \cite{lee20232} first leverages the monocular depth estimation algorithm \cite{watson2021temporal} to recover 3D information from RGB images before cross-modal matching. \mbox{I2P-Rec} \cite{zheng2023i2p} also converts RGB images to depth maps, but it then transforms both the depth maps and point clouds into unified bird's-eye views. Another line is to implicitly align modalities by learning a shared 2D-3D embedding space. For example, the pioneering work \cite{cattaneo2020global} adopts a teacher-student paradigm to jointly learn descriptors for images and point clouds. ModalLink \cite{xie2024modalink} designs a non-negative factorization-based encoder to generate descriptors. VXP \cite{li2024vxp} voxelizes point clouds and enhances global descriptor consistency by establishing correspondences between voxels and pixels. However, these methods either overlook the FOV differences \cite{cattaneo2020global, lee20232, li2024vxp}, sacrificing the quality of descriptors, or crop the point cloud’s FOV to align with RGB images \cite{zheng2023i2p, xie2024modalink}, limiting practical application. Superior to them, our method takes 360° point clouds as input, efficiently generating dense descriptors for multiple viewpoints in a single pass and improving retrieval accuracy through multi-view matching.
 

\subsection{State Space Model and Its Evolution}
State-space model (SSM) originally stemmed from the Kalman filter \cite{kalman1960new}, designed to describe the input-output relationships in dynamic systems. Recently, it has evolved into a deep learning tool suited for sequence modeling. LSSL \cite{gu2021combining} introduces HiPPO \cite{gu2020hippo} to initialize SSM parameters and demonstrates its capability to handle long-range dependencies. The Structured SSM (S4) \cite{gu2022efficiently} normalizes the parameter matrices into a diagonal structure and achieves fast generative modeling and regression. Mamba \cite{gu2023mamba} proposes a novel selective scanning mechanism to dynamically adjust SSM parameters based on input, controlling information propagation and filtering in a way distinct from attention. Besides, Mamba also provides a hardware-aware parallel algorithm for fast convolution \cite{gu2022efficiently, gu2023mamba} in a recurrent mode. These features enable SSM to efficiently capture long-range relationships with linear complexity, making it a potential alternative to Transformers in language \cite{mehta2022long} and audio \cite{shams2024ssamba, yadav2024audio} tasks. Subsequently, visual state space model (VMamba) has emerged as a generic backbone for image processing \cite{liu2024vmamba, zhu2024vision}. The recent 2D vision applications include classification \cite{li2024mambahsi, yue2024medmamba}, object detection \cite{chen2024mim, wang2024mamba} and semantic segmentation \cite{wang2024large, ma2024rs}. Additionally, some researchers are now extending VMamba to the field of 3D vision, such as VoxelMamba \cite{zhang2024voxel} and PointMamba \cite{liu2024point}. Inspired by the success of these applications, we also take VMamba as the backbone to generate descriptors for place recognition. Notably, the input of our model is multimodal data, including RGB images and point clouds. 


\section{Methodology}
This section begins with the principles of visual state space model (VMamba), then details our proposed method, including the backbone, descriptor generation with improved NetVLAD, the definition of cross-modal similarity, pixel-view-scene joint contrastive learning.

\begin{figure}[t]
    \centering
    \includegraphics[width=1\linewidth]{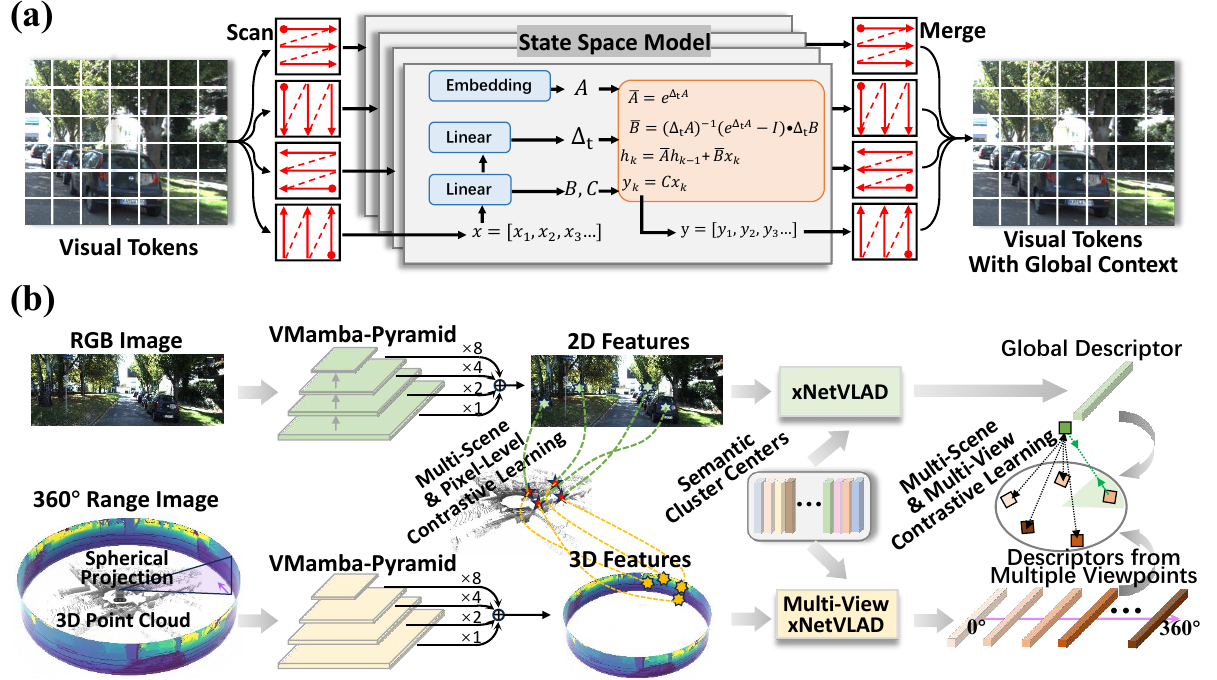}
    \vspace{-5mm}
    \caption{Illustrations of VMamba and our cross-modal framework. (a) VMamba converts 2D visual tokens to 1D sequences for the state space model using a four-way scanning mechanism. (b) Our framework builds VMamba-based Pyramids for RGB images and point clouds, respectively. Point clouds are converted to 360° range images via spherical projection, offering a unified 2D format. After extracting pixel features from both modalities, the improved NetVLAD (denoted as xNetVLAD) generates a global descriptor for RGB images, and efficiently generates multi-view descriptors for 360° range images. We perform pixel-view-scene joint contrastive learning to train the dual-pyramid model.}
    \label{fig2}
    \vspace{-3mm}
\end{figure}

\subsection{VMamba-based Backbone}
\textbf{Preliminaries}. SSM is a mathematical model in Kalman filter \cite{kalman1960new} to define linear time-invariant systems, which describes the relationship between continuous input $x(t) \in \mathbb{R}$ and output $y(t) \in \mathbb{R}$ through a hidden state $h(t) \in \mathbb{R}^{N}$ as:
\begin{equation}
\begin{aligned}
    \label{eq:ssm1}
    h'(t) &= \mathbf{A} h(t) +  \mathbf{B} x(t),\\
    y(t) &= \mathbf{C} h(t).
\end{aligned}
\end{equation}
Here, $\mathbf{A} \! \in \! \mathbb{R}^{N \times N}$ is a state transition matrix while $\mathbf{B} \! \in \! \mathbb{R}^{N \times 1}$ and $\mathbf{C} \in \mathbb{R}^{N \times 1}$ are the projection matrices. To be integrated into deep learning, some prior works \cite{gu2022efficiently, gu2023mamba} discretize the above system with a timescale parameter $\Delta_t$ as:
\begin{equation}
\begin{aligned}
    \label{eq:ssm2}
    \bar{\mathbf{A}} &= e^{\Delta_t \mathbf{A}},\\
    \bar{\mathbf{B}} &= {(\Delta_t \mathbf{A})}^{-1}(e^{\Delta_t \mathbf{A}} - \mathbf{I}) \cdot \Delta_t \mathbf{B},\\
    h_k &= \bar{\mathbf{A}} h_{k-1} +  \bar{\mathbf{B}} x_k,\\
    y_k &= \mathbf{C} h_k,
\end{aligned}
\end{equation}
where $k$ represents the discrete index. Besides, a selective scanning algorithm \cite{gu2023mamba} is designed to associate the time-invariant parameters with inputs, equipped with a global convolution algorithm \cite{martin2017parallelizing, smith2022simplified} for parallel acceleration. Based on these, SSM has achieved information interaction similar to attention mechanism, but with linear complexity. To extend its visual application, VMamba \cite{liu2024vmamba} designs a four-way scanning strategy to organize visual tokens into different 1D sequences and then merges the multi-path SSM features to capture global context, as shown in \mbox{Fig. \ref{fig2} (a)}.

\textbf{Cross-Modal Backbone.} We implemented two separate VMamba-based pyramids to extract features from RGB images and point clouds, as shown in \mbox{Fig. \ref{fig2} (b)}. To obtain a unified data format, point clouds are converted into 360° range images via spherical projection \cite{wu2018squeezeseg}, which also facilitates the later multi-view processing. Inside each pyramid, we repeatedly apply the forward process of VMamba at various image scales, and then concatenate the multi-scale SSM features for fusion. Besides, NetVLAD was upgraded to generate descriptors for both modalities.

\subsection{Global Descriptors of RGB Images}
\label{sec:netvlad}
\textbf{Generation.} Given an RGB image $\boldsymbol{I} \! \in \! \mathbb{R}^{\bar{H} \! \times \! \bar{W} \! \times \! 3}$, the VMamba-based backbone extracts features $ f^{2d} \! \in \! \mathbb{R}^{\bar{H} \! \times \! \bar{W} \! \times \! d_f} $, where $d_f$ is the feature dimension. Following the mainstream pipeline \cite{ZHANG2021107952}, $f^{2d}$ is aggregated into a global descriptor $\mathcal{G}^{2d} \! \in \! \mathbb{R}^{d_g}$ via NetVLAD \cite{arandjelovic2016netvlad}. It learns $K$ semantic cluster centers \mbox{$\{c_1,...,c_{K} \,|\, c_k \! \in \! \mathbb{R}^{d_f}\}$}, and then computes a global semantic vector $ \boldsymbol{F}^{2d}_k \! \in \! \mathbb{R}^{d_f} $ for each center $c_k$ by pixel-wise residual summation as:
\begin{align}
    \label{eq3:residual}
    \boldsymbol{F}^{2d}_k = \sum_{i=1}^{\bar{H}}\sum_{j=1}^{\bar{W}} \gamma_k(f^{2d}_{ij})\cdot(f^{2d}_{ij} - c_k),
\end{align}
where $f^{2d}_{ij}$ is the feature vector $f^{2d}[i,j,:]$ at pixel ($i,j$), and $\gamma_k$($\cdot$) computes the soft-assignment weights via MLP and softmax. Finally, we get a ($K\!\times\!\,d_f$)-dimensional semantic vector and compress it into $\mathcal{G}^{2d}$ using another MLP.

\begin{figure}[t]
    \centering
    \includegraphics[width=1\linewidth]{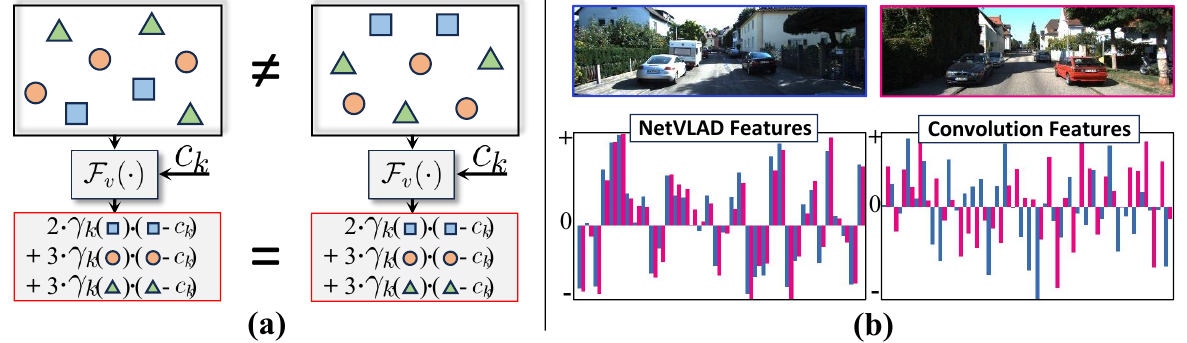}
    \vspace{-7mm}
    \caption{Limitations of vanilla NetVLAD. (a) A toy example to demonstrate the geometric information loss. Colorful shapes represent different semantic objects. The residual summation operator $\mathcal{F}_{v}$ in Eq. \ref{eq3:residual} generates the same feature vector for the two scenes with same semantics while different geometries. (b) Two real-world scenes with similar semantics, both containing road, cars, buildings, and vegetation. The bottom two figures are the features generated by NetVLAD and convolution. Blue and red represent different scenes. The x-axis and y-axis represent indices and values, respectively.}
    \label{fig3}
    \vspace{-3mm}
\end{figure}

\textbf{Geometric Compensation.} We observed that the vanilla NetVLAD often retrieves incorrect scenes with semantics similar to the target. This is because the residual summation fails to capture the geometric distribution of semantic objects, as shown in Fig. \ref{fig3} (a). However, this distribution is crucial for outdoor place recognition, as most scenes contain similar semantic objects, such as a road flanked by many trees or buildings. Therefore, we designed an additional convolution branch inside NetVLAD for geometric compensation, since 2D kernels can easily capture geometric differences, as shown in Fig. \ref{fig3} (b). The upgraded NetVLAD (denoted as xNetVLAD) take an MLP to merge the semantic and geometric features into a global descriptor $\mathcal{G}^{2d}$ as:
\begin{align}
    \label{eq4:xnetvlad}
    \mathcal{G}^{2d} = \mathrm{MLP}\left(\boldsymbol{F}^{2d}_1,\dots,\boldsymbol{F}^{2d}_K, \mathrm{Conv}(f^{2d})\right).
\end{align}

\subsection{Multi-View Descriptors of Point Clouds}
\label{sec:mvg}
\textbf{Multi-View Generation.} Given a point cloud $\boldsymbol{P}$, we feed its 360° range image $\boldsymbol{D} \! \in \! \mathbb{R}^{\tilde{H} \! \times \! \tilde{W}}$ into the VMamba-based backbone and get $ f^{3d} \! \in \! \mathbb{R}^{\tilde{H} \times \tilde{W} \times d_f} $, where $\tilde{H}$ and $\tilde{W}$ represent the vertical and horizontal angle resolutions of spherical projection \cite{wu2018squeezeseg}. Generally, the FOV of camera is much smaller than that of LiDAR, so the global descriptors of point clouds and RGB images can not be perfectly aligned in terms of scene content. To address it, an intuitive approach is to slice $\boldsymbol{D}$ from $n_v$ horizontal viewpoints as \mbox{$\{v_1,...,v_{n_v} \,|\, v_j \! \in \! \mathbb{R}^{\tilde{H} \times L}, L < \tilde{W}\}$}, where $L$ is the view width, $v_j \! = \! \boldsymbol{D}[:,\,\Delta \! \cdot \! j:\Delta \! \cdot \! j \! + \! L]$ and $\Delta$ is the view offset. Then, the multi-view descriptors are generated as:
\begin{align}
    \label{eq5:slice}
    \mathcal{G}^{3d}_{v_j} = \mathrm{xNetVLAD}(f^{3d}_{v_j}),
\end{align}
where $f^{3d}_{v_j}$ is the corresponding features at the view $v_j$.

\textbf{Efficient Inference.} Eq. \ref{eq5:slice} allocates memory and runs inference for each view from scratch. To avoid missing some key views, we usually set $\Delta \! < \! L$, causing overlap between adjacent views. As $\Delta$ decreases, the probability of accurately capturing the RGB image scenes increases, but the required resources also increase. To address it, our key insight is to defer the slicing operation. Since slicing is only along the horizontal direction, the residual summation in \mbox{Eq. \ref{eq3:residual}} can be pre-computed vertically as:
\begin{align}
    \label{eq6:vertical}
    \mathcal{V}^{k}_j = \begin{matrix}\sum_{i=1}^{\bar{H}}\gamma'_k(f^{3d}_{ij})\cdot(f^{3d}_{ij} - c_k)\end{matrix},
\end{align}
where $1\!<j\!<\tilde{W}$ and $\mathcal{V}^{k} \! \in \! \mathbb{R}^{\tilde{W} \times d_f}$. Then, slicing and residual aggregation along the $\tilde{W}$-dimension can save many resources. Note that the convolutional branch can also perform vertical pooling first. The final descriptor of each view is consistent with \mbox{Eq. \ref{eq5:slice}}.

\subsection{Similarity between RGB Images and Point Cloud Views}


We train on a dataset of RGB images and point clouds, captured synchronously along the motion trajectory. At time step $t$, we denote the camera-centered coordinate frame as $C_t$ and the LiDAR-centered coordinate frame as $L_t$. Typically, the camera poses of the trajectory are known, which transform the coordinate frame $C_t$ to the world coordinate frame $W$ with $\mathbf{T}_{WC_t} \! \in \! \mathbb{R}^{4 \times 4}$. The camera to LiDAR coordinate transformation is invariant and calibrated as $\mathbf{T}_{LC} \! \in \! \mathbb{R}^{4 \times 4}$. 

\begin{figure}[t]
    \centering
    \includegraphics[width=1\linewidth]{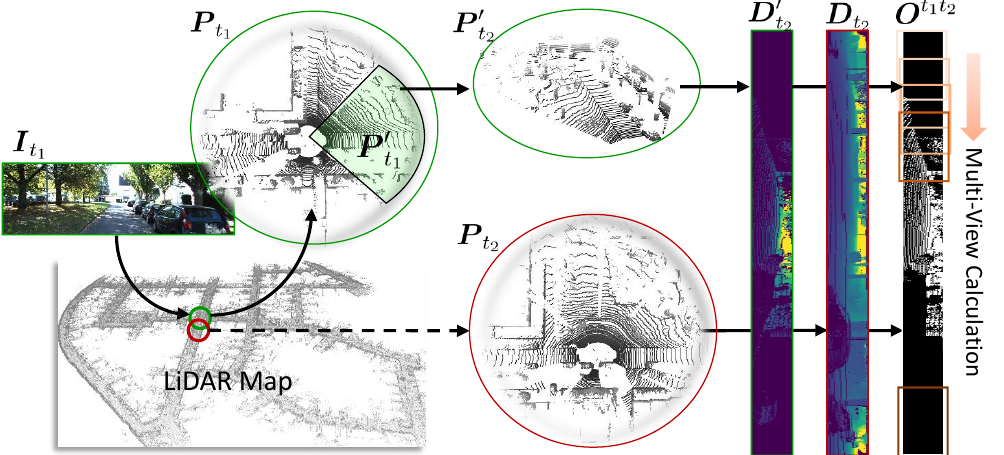}
    \vspace{-7mm}
    \caption{The workflow to quantify the similarity between RGB image $\boldsymbol{I}_{t_1}$ and the multiple views of point cloud $\boldsymbol{P}_{t_2}$. We use it solely for training.}
    \label{fig4}
    \vspace{-3mm}
\end{figure}

Given an RGB image $\boldsymbol{I}_{t_1}$ and a point cloud $\boldsymbol{P}_{t_2}$, the modality gaps make it difficult to quantify the multi-view similarity. Our key insight is to extract the 3D points from LiDAR map for the RGB image, and then calculate the point overlap \cite{chen2022overlapnet}. As illustrated in Fig. \ref{fig4}, we first find the point cloud $\boldsymbol{P}_{t_1}$ synchronized with $\boldsymbol{I}_{t_1}$, as they observe the same scene surface. The subset $\boldsymbol{P}'_{t_1}$ is selected from $\boldsymbol{P}_{t_1}$ as:
\begin{align}
    \boldsymbol{P}'_{t_1} = \mathcal{F}_{p}(\mathbf{T}_{LC}^{-1}\boldsymbol{P}_{t_1}),
\end{align}
where $\mathcal{F}_{p}$($\cdot$) filters out the 3D points outside the camera FOV, and $\boldsymbol{P}'_{t_1}$ is in the coordinate frame $C_{t_1}$. Then, we reproject $\boldsymbol{P}'_{t_1}$ into the coordinate frame $L_{t_2}$ as:
\begin{align}
    \boldsymbol{P}'_{t_2} = \mathbf{T}_{LC} \mathbf{T}_{WC_{t_2}}^{-1} \mathbf{T}_{WC_{t_1}} \boldsymbol{P}'_{t_1}.
\end{align}
Using spherical projection, we generate 360° range images $\boldsymbol{D}_{t_2} \! \in \! \mathbb{R}^{\tilde{H} \! \times \! \tilde{W}}$ and $\boldsymbol{D}'_{t_2} \! \in \! \mathbb{R}^{\tilde{H} \! \times \! \tilde{W}}$ for $\boldsymbol{P}_{t_2}$ and $\boldsymbol{P}'_{t_2}$. Afterwards, the binary visible map $\boldsymbol{O}^{t_1t_2} \! \in \! \mathbb{R}^{\tilde{H} \! \times \! \tilde{W}}$ is defined as:
\begin{align}
    \boldsymbol{O}^{t_1t_2}_{ij} \! = \! \llbracket \boldsymbol{D}'_{t_2}[i,j] > 0 \land \left| \boldsymbol{D}_{t_2}[i,j] \! - \!\boldsymbol{D}'_{t_2}[i,j]  \right| < \epsilon \rrbracket,
\end{align}
where $\llbracket\cdot\rrbracket$ is the Iversion bracket, $\boldsymbol{D}'_{t_2}[i,j] > 0$ indicates that at least one 3D point is projected onto pixel ($i,j$), and $\epsilon$ is a distance threshold. We consider the corresponding 3D point in $\boldsymbol{I}_{t_1}$ as visible in $\boldsymbol{P}_{t_2}$ only when $\left| \boldsymbol{D}_{t_2}[i,j] \! - \!\boldsymbol{D}'_{t_2}[i,j]  \right| < \epsilon$. Following section \ref{sec:mvg}, we slice $\boldsymbol{O}^{t_1t_2}$ into multiple views as $\{\boldsymbol{O}^{t_1t_2}_{v_1}, \dots, \boldsymbol{O}^{t_1t_2}_{v_{n_v}} | \boldsymbol{O}^{t_1t_2}_{v_j} \! \in \! \mathbb{R}^{\tilde{H} \times L} \}$. The similarity $\mathcal{S}^{t_1t_2}_{v_j} \in \mathbb{R}$ between $\boldsymbol{I}_{t_1}$ and the view $v_j$ of $\boldsymbol{D}_{t_2}$ is defined as:
\begin{align}
    \label{eq9:similarity}
    \mathcal{S}^{t_1t_2}_{v_j} = \frac{\sum_{(i',j')} \boldsymbol{O}^{t_1t_2}_{v_j}[i',j'] }{\lVert \boldsymbol{P}'_{t_1} \rVert},
\end{align}
where $0 \! \leq \! \mathcal{S}^{t_1t_2}_{v_j} \! \leq \! 1$, and $\lVert \boldsymbol{P}'_{t_1} \rVert$ is the point number in $\boldsymbol{P}'_{t_1}$.

\subsection{Pixel-View-Scene Joint Contrastive Learning}
The circle loss \cite{sun2020circle} is adopted for joint contrastive learning across pixels, views, and scenes. Given a template anchor $s_a$, its positive sample set is denoted as $\Omega_{+} \! = \! \{s^1_+, \dots, s^{n^+}_+\}$, and its negative sample set is denoted as $\Omega_{-} \! = \! \{s^1_-, \dots, s^{n^-}_-\}$. The template loss function $\mathcal{F}_{c}(s_a)$ is formulated as: 
\begin{align}
    \label{eq10:circle}
    \mathcal{F}_{c}(s_a) = \log \! \left( 1 \! + \! \sum_{j\in\Omega_+} e^{\theta_+^j(d^j_+ - \Delta_+)} \cdot \! \sum_{k\in\Omega_{-}} e^{\theta_-^k(\Delta_- - d_-^k)} \right),
\end{align}
where $d^j_+$ is the feature distance between $s_a$ and the positive sample $s^j_+$, $d^k_-$ is the feature distance between $s_a$ and the negative sample $s^k_-$, $\Delta_+$ and $\Delta_-$ are the positive and negative margins, $\theta_+^j = \lambda(d_+^j \! - \! \Delta_+)$ and $\theta_-^k = \lambda(\Delta_- \! - \! d_-^k)$ are the adaptive weights with $\lambda$ being a scaling factor.

\textbf{Pixel-Level Loss.} Given $\boldsymbol{I}_{t_1}$ and $\boldsymbol{P}_{t_2}$. we convert each point $\mathbf{p}_i \in \boldsymbol{P}_{t_2}$ to the pixel coordinates ($\bar{u}_i, \bar{v}_i$) of $\boldsymbol{I}_{t_1}$ via pin-hole projection, and to the pixel coordinates ($\tilde{u}_i, \tilde{v}_i$) of $\boldsymbol{D}_{t_2}$ via spherical projection. This bidirectional relationship is used to index cross-modal samples. Taking the 2D features $f^{2d}_{t_1}[\bar{u}_i, \bar{v}_i]$ of $\boldsymbol{I}_{t_1}$ at pixel ($\bar{u}_i, \bar{v}_i$) as the anchor, its cross-modal positive sample set $\Omega_{+}^{3D}(t_1 \! \to \! t_2)$ and the cross-modal negative set $\Omega_{-}^{3D}(t_1 \! \to \! t_2)$ are collected from the 3D features $f^{3d}_{t_2}$ of $\boldsymbol{D}_{t_2}$. If the pixel ($\bar{u}_j, \bar{v}_j$) is close enough to ($\bar{u}_i, \bar{v}_i$), we define $f^{3d}_{t_2}[\tilde{u}_j, \tilde{v}_j]$ as a positive sample of $f^{2d}_{t_1}[\bar{u}_i, \bar{v}_i]$; otherwise, it is considered a negative sample. We will select $n_s$ pixels from $\boldsymbol{I}_{t_1}$ to calculate the average loss as:
\begin{align}
    \label{eq11:pixel}
    \mathcal{L}_{p} = \frac{1}{n_s} \sum_{i=1}^{n_s} \left( \,\,\mathcal{F}_{c}\left(f^{2d}_{t_1}[\bar{u}_i, \bar{v}_i]\right)\,\,\right).
\end{align}

\begin{table*}[t]
\centering
\scriptsize
\caption{Recall@N (\%) on the KITTI dataset. The best results in each column are highlighted in bold, and the seconds are underlined.}
\vspace{-2mm}
\label{tab:recall-kitti}
\renewcommand\arraystretch{1.1}
\begin{tabular}{c|c|c|c|c|c|c}
\hline
           & KITTI-00                & KITTI-02                & KITTI-05                & KITTI-06                & KITTI-07                & KITTI-08                \\ \hline
Method     & R@1\,/\,R@5\,/\,R@1\%     & R@1\,/\,R@5\,/\,R@1\%     & R@1\,/\,R@5\,/\,R@1\%     & R@1\,/\,R@5\,/\,R@1\%     & R@1\,/\,R@5\,/\,R@1\%     & R@1\,/\,\,R@5\,/\,\,R@1\%     \\ \hline
PlainEBD\cite{cattaneo2020global} & 19.93\,\,/\,\,31.71\,\,/\,\,69.06 & 16.50\,\,/\,\,27.48\,\,/\,\,55.93 & 30.64\,\,/\,\,46.11\,\,/\,\,71.10 & 25.34\,\,/\,\,35.88\,\,/\,\,47.68 & 39.95\,\,/\,\,59.30\,\,/\,\,70.93 & 20.49\,\,/\,\,35.96\,\,/\,\,72.22 \\
(LC)$^2$ \cite{lee20232}        & 30.65\,\,/\,\,47.76\,\,/\,\,84.52 & 23.06\,\,/\,\,35.72\,\,/\,\,67.41 & 40.49\,\,/\,\,54.58\,\,/\,\,80.62 & 39.78\,\,/\,\,58.22\,\,/\,\,71.48 & 52.50\,\,/\,\,68.03\,\,/\,\,84.01 & 38.42\,\,/\,\,54.07\,\,/\,\,87.30 \\
I2P-Rec \cite{zheng2023i2p} \textcolor{red}{*}  & 44.84\,\,/\,\,59.55\,\,/\,\,90.09 & 28.00\,\,/\,\,42.33\,\,/\,\,71.68 & 48.32\,\,/\,\,64.61\,\,/\,\,88.52 & 43.51\,\,/\,\,63.12\,\,/\,\,72.30 & 63.03\,\,/\,\,73.39\,\,/\,\,78.93 & 45.91\,\,/\,\,62.44\,\,/\,\,91.65 \\
VXP \cite{li2024vxp}        & 24.22\,\,/\,\,38.16\,\,/\,\,80.00 & 17.72\,\,/\,\,30.83\,\,/\,\,64.64 & 32.81\,\,/\,\,50.63\,\,/\,\,78.16 & 29.97\,\,/\,\,41.96\,\,/\,\,52.50 & 43.69\,\,/\,\,61.31\,\,/\,\,76.29 & 24.01\,\,/\,\,37.68\,\,/\,\,79.17 \\
ModaLink \cite{xie2024modalink} \textcolor{red}{*}        & \underline{90.31}\,\,/\,\,\underline{95.53}\,\,/\,\,\underline{99.58} & \underline{70.18}\,\,/\,\,\underline{81.85}\,\,/\,\,\underline{96.74} & \underline{90.00}\,\,/\,\,\underline{95.00}\,\,/\,\,\textbf{99.31} & \underline{86.19}\,\,/\,\,\underline{96.64}\,\,/\,\,\textbf{100.0} & \underline{96.55}\,\,/\,\,\underline{99.27}\,\,/\,\,\textbf{100.0} & \underline{86.97}\,\,/\,\,\textbf{95.75}\,\,/\,\,\textbf{99.93} \\
Ours       & \textbf{93.13}\,\,/\,\,\textbf{96.83}\,\,/\,\,\textbf{99.74} & \textbf{76.98}\,\,/\,\,\textbf{86.44}\,\,/\,\,\textbf{97.51} & \textbf{90.76}\,\,/\,\,\textbf{96.09}\,\,/\,\,\underline{99.24} & \textbf{89.28}\,\,/\,\,\textbf{97.73}\,\,/\,\,\underline{99.18} & \textbf{98.09}\,\,/\,\,\textbf{99.46}\,\,/\,\,\underline{99.82} & \textbf{90.64}\,\,/\,\,\underline{95.23}\,\,/\,\,\underline{99.63} \\ \hline 
\end{tabular}
\footnotesize{The methods marked with \textcolor{red}{*} need cropping point clouds, while others cover the full 360° view.\qquad\qquad\qquad\qquad\qquad\qquad\qquad\qquad\qquad\qquad\quad}
\vspace{-3mm}
\end{table*}

\textbf{Multi-View Loss.} We use Eq. \ref{eq9:similarity} only for creating multi-view supervision. For the global descriptor $\mathcal{G}^{2d}(t_1)$ of $\boldsymbol{I}_{t_1}$ and the multi-view 3D descriptors $\{ \mathcal{G}^{3d}_{v_1}(t_2),\dots,\mathcal{G}^{3d}_{v_{n_v}}(t_2)\}$ of $\boldsymbol{D}_{t_2}$, the similarity scores are $\{\mathcal{S}^{t_1t_2}_{v_1},\dots, \mathcal{S}^{t_1t_2}_{v_{n_v}} \}$. Taking $\mathcal{G}^{2d}(t_1)$ as the anchor, the positive sample set $\Psi^{3d}_{+}(t_1 \! \to \! t_2)$ contains the 3D descriptors with the similarity scores above $\tau^+_s$, while the negative set $\Psi^{3d}_{-}(t_1 \! \to \! t_2)$ contains those with scores below $\tau^-_s$. The multi-view loss from $\boldsymbol{I}_{t_1}$ to $\boldsymbol{D}_{t_2}$ is:
\begin{align}
    \label{eq12:view}
    \mathcal{L}_{v} = \mathcal{F}_{c}\left( \,\,\mathcal{G}^{2d}(t_1) \,\,\right).
\end{align}

\textbf{Multi-Scene Loss.} $\boldsymbol{I}_{t_1}$ and $\boldsymbol{P}_{t_2}$ are considered positively correlated if the distance between poses $\mathbf{T}_{WC_{t_1}}$ and $\mathbf{T}_{WC_{t_2}}$ is less than $\tau_d^+$. The pixel-level and multi-view loss can only be computed under this condition, since a large distance, i.e., exceeding $\tau_d^-$, results in empty positive sample sets in Eq. \ref{eq11:pixel} and \ref{eq12:view}. To mine knowledge from distant scenes, we introduce a random negative point cloud $\boldsymbol{P}_{t_3}$ here. We collect its 3D features from $f^{3d}_{t_3}$ and add them to the negative sample set $\Omega_{-}^{3D}(t_1 \! \to \! t_2)$ to form the extended set $\Omega_{-}^{3D}(t_1 \! \to \! t_2, t_3)$. Based on it, the extended pixel-level loss is denoted as $\mathcal{L}_{p}^{ms}$.  The negative sample set $\Psi^{3d}_{-}(t_1 \! \to \! t_2,t_3)$ is generated in a similar way to compute the extended multi-view loss as $\mathcal{L}_{v}^{ms}$. 

Finally, the pixel-view-scene joint loss is formulated as:
\begin{align}
    \label{eq15:ms}
    \mathcal{L}_{joint} = \mathcal{L}_{p}^{ms} + \mathcal{L}_{v}^{ms}.
\end{align}


\begin{figure}[t]
\centering
\includegraphics[width=0.48\textwidth]{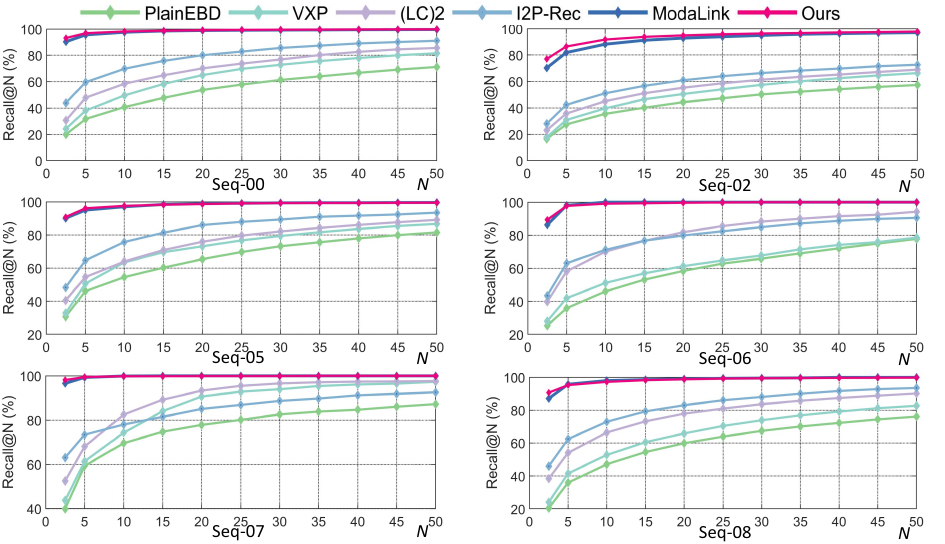} 
\vspace{-7mm}
\caption{The Recall@N (ranging from 0 to 50) results on KITTI.}
\label{fig5:recall}
\vspace{-3mm}
\end{figure}

\section{Experiments}
\subsection{Experimental Settings}
\textbf{Implementation Details.} The experiments are conducted on an Intel i7-12700K CPU and NVIDIA RTX 3090 GPU. We follow this work \cite{liu2024vmamba} to implement our VMamba-based pyramid with PyTorch 2.1.1, and train it using the AdamW optimizer with an initial learning rate of $1e^{-4}$ decaying by 0.8 every 5 epochs. For spherical projection, the resolution of 360° range image is $48\times900$, with pixels that have no 3D point projections set to zero. Since the width of RGB camera FOV approximates 200 pixels in the range images, $L$ is set to 200 for multi-view generation, and the view offset $\Delta$ is set to 10 and 30 pixels for training and testing. Due to GPU memory limits, the feature dimension $d_f$, number of cluster center $K$, and descriptor dimension $d_g$ are set to 64, 48, and 256, respectively. For similarity calculation, $\epsilon$ is set to 1m to determine the visibility of 3D points. For pixel-level loss, we sample $n_s\!=\!512$ anchor pixels, and set the margins $\Delta_+$ and $\Delta_-$ to 0.1 and 1.4. For multi-view loss, we set the similarity thresholds $\tau^+_s$ and $\tau^-_s$ to 0.6 and 0.2, and set the margins $\Delta_+$ and $\Delta_-$ to 0.4 and 1.2. For multi-scene loss, we set the distance thresholds $\tau_d^+$ and $\tau_d^-$ to 3m and 20m.

\textbf{Datasets.} Two real-world outdoor datasets, i.e., KITTI \cite{geiger2013vision} and KITTI-360 \cite{liao2022kitti} are used for experiments. Both datasets include the calibration data, system trajectories, and synchronized RGB images and point clouds. The KITTI dataset contains 22 sequences, with the last 11 for training and the first 11 for testing. Although only sequences 00 to 11 contain the original ground-truth poses, we obtain the poses of other sequences from SemanticKITTI \cite{semantickitti}. In contrast, KITTI-360 contains 11 longer sequences (about 73.7 KM), making it more challenging for place recognition.

\begin{table}[t]
\centering
\caption{Recall@N (\%) on the KITTI dataset. The point clouds are rotated using a random yaw angle within a 360° range.}
\vspace{-2mm}
\label{tab2:yaw}
\begin{tabular}{c|c|c}
\hline
           & KITTI-00                & KITTI-02                \\ \hline
Method     & R@1\,\,/\,\,R@5\,\,/\,\,R@1\%     & R@1\,\,/\,\,R@5\,\,/\,\,R@1\%     \\ \hline
PlainEBD\cite{cattaneo2020global} & 13.51\,\,/\,\,26.65\,\,/\,\,58.73 & 9.79\,\,/\,\,23.37\,\,/\,\,49.43 \\
(LC)$^2$ \cite{lee20232}        & \underline{26.13}\,\,/\,\,\underline{44.21}\,\,/\,\,\underline{82.55} & \underline{21.46}\,\,/\,\,\underline{31.27}\,\,/\,\,\underline{62.48} \\ 
I2P-Rec \cite{zheng2023i2p} \textcolor{red}{*} & 18.27\,\,/\,\,24.43\,\,/\,\,37.17 & 11.89\,\,/\,\,16.97\,\,/\,\,28.35 \\ 
VXP \cite{li2024vxp}        & 17.57\,\,/\,\,31.41\,\,/\,\,71.12 & 14.19\,\,/\,\,24.23\,\,/\,\,59.62 \\ 
ModaLink \cite{xie2024modalink} \textcolor{red}{*}   & 15.54\,\,/\,\,16.97\,\,/\,\,19.57 & 13.07\,\,/\,\,14.76\,\,/\,\,18.85 \\
Ours       & \textbf{93.22}\,\,/\,\,\textbf{97.01}\,\,/\,\,\textbf{99.80} & \textbf{76.76}\,\,/\,\,\textbf{86.57}\,\,/\,\,\textbf{97.64} \\ \hline
\end{tabular}
\vspace{-3mm}
\end{table}

\textbf{Evaluation Protocol.} Image-to-point cloud place recognition targets camera localization within a pre-built LiDAR map. In each dataset, all point clouds form the LiDAR map, while monocular RGB images serve as online queries. In real-world applications, the camera can not revisit the exact same location where the point clouds were captured. To account for such positional offsets, we will remove the point cloud with the same frame ID as each image query.


\subsection{Results of Image-to-Point Cloud Place Recognition}

\textbf{Baselines.} We compare our method with others, including the basic 2D-3D embeddings \cite{cattaneo2020global} (denoted as PlainEBD), (LC)$^2$ \cite{lee20232}, the monocular version of I2P-Rec \cite{zheng2023i2p}, VXP \cite{li2024vxp} and ModaLink \cite{xie2024modalink}. PlainEBD and VXP are reproduced using this 3D sparse convolution framework \cite{yan20222dpass}. I2P-Rec is reproduced using ResNet-34 and monocular depth estimation MIM-Depth \cite{xie2023revealing}. (LC)$^2$ and ModaLink are open-sourced but retrained. Note that some methods use camera pose priors to crop point clouds, i.e., I2P-Rec keeps only the front half, and ModaLink keeps only the points within the camera FOV. In localization applications, cropping is not feasible since the camera poses relative to the LiDAR map are unknown. 

\begin{table}[t]
\centering
\caption{Recall@N (\%) on the KITTI-360 dataset of our method.}
\vspace{-2mm}
\label{tab3:kitti360}
\begin{tabular}{c|c|c}
\hline
     & KITTI-360 Full Length             & KITTI-360 1/3 Length              \\ \hline
Seq  & R@1\,/\,R@5\,/\,R@10\,/\,R@1\%      & R@1\,/\,R@5\,/\,R@10\,/\,R@1\%      \\ \hline
0000 & 51.94\,\,/\,\,67.36\,\,/\,\,76.18\,\,/\,\,96.93 & 64.16\,\,/\,\,76.80\,\,/\,\,84.70\,\,/\,\,94.92 \\ 
0002 & 39.56\,\,/\,\,53.64\,\,/\,\,62.97\,\,/\,\,94.79 & 54.58\,\,/\,\,66.18\,\,/\,\,74.56\,\,/\,\,93.13 \\ 
0004 & 47.54\,\,/\,\,61.15\,\,/\,\,70.30\,\,/\,\,94.04 & 57.57\,\,/\,\,71.44\,\,/\,\,80.32\,\,/\,\,93.57 \\ 
0005 & 53.87\,\,/\,\,66.62\,\,/\,\,75.71\,\,/\,\,93.56 & 67.48\,\,/\,\,76.82\,\,/\,\,84.36\,\,/\,\,90.03 \\
0006 & 54.34\,\,/\,\,68.64\,\,/\,\,76.94\,\,/\,\,95.92 & 71.39\,\,/\,\,82.63\,\,/\,\,88.63\,\,/\,\,96.34 \\ 
0009 & 54.53\,\,/\,\,69.41\,\,/\,\,77.37\,\,/\,\,97.84 & 64.94\,\,/\,\,78.21\,\,/\,\,86.30\,\,/\,\,97.37 \\ \hline
\end{tabular}
\end{table}

\begin{figure}[t]
\centering
\includegraphics[width=0.48\textwidth]{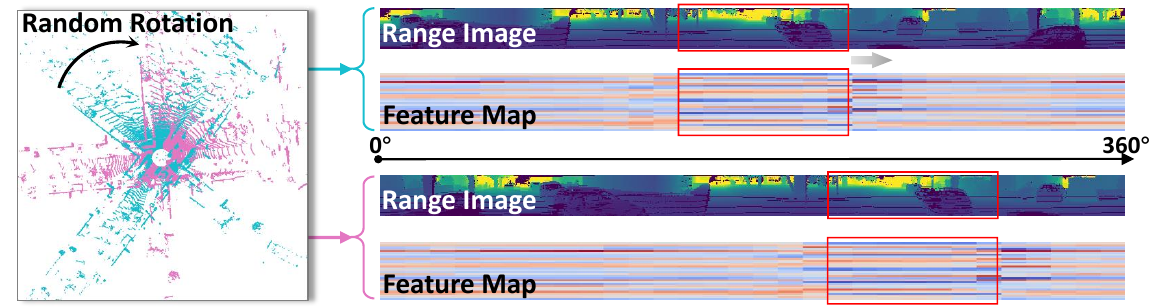} 
\vspace{-6mm}
\caption{Yaw rotation only shifts the feature map horizontally, so it does not change the multi-view descriptors but changes their generation order.}
\label{fig6:rotation}
\vspace{-3mm}
\end{figure}

\begin{table}[b]
\centering
\vspace{-3mm}
\caption{Ablation studies on KITTI-00 for Loss and xNetVLAD.}
\vspace{-2mm}
\label{tab4:loss_netvlad}
\begin{tabular}{c|c|c|c|c|c}
\hline
View & Scene & Pixel & GEOC & EFFI & R@1\,/\,R@5\,/\,R@1\% \\ \hline
$\checkmark$  &        & $\checkmark$  & $\checkmark$  & $\checkmark$  & 14.58\,\,/\,\,31.89\,\,/\,\,71.66                   \\  
$\checkmark$  & $\checkmark$  &        & $\checkmark$  & $\checkmark$  & 46.53\,\,/\,\,63.14\,\,/\,\,90.49                 \\  
$\checkmark$  & $\checkmark$  & $\checkmark$  &        & $\checkmark$  &       73.71\,\,/\,\,85.27\,\,/\,\,98.59             \\  
$\checkmark$  & $\checkmark$  & $\checkmark$  & $\checkmark$  &        &   N/A\,\,\,/\,\,\,N/A\,\,\,/\,\,\,N/A                 \\  
$\checkmark$  & $\checkmark$  & $\checkmark$  & $\checkmark$  & $\checkmark$ &      93.13\,\,/\,\,96.83\,\,/\,\,99.74        \\  \hline
\end{tabular}
\footnotesize{Without EFFI, we encountered \textcolor{red}{out-of-memory} errors during training.}
\vspace{-3mm}
\end{table}

\textbf{Train and test on KITTI.} We quantify the place recognition performance using Recall@N (\%), which denotes the proportion of successful recognitions. For each query RGB image, we retrieve N point clouds as candidates, and deem this recognition successful if any is less than 10m from the image's location. For our multi-view matching, if any view in a point cloud matches the RGB image, the point cloud is added to the candidate set. We report the statistical results in \mbox{Table \ref{tab:recall-kitti}}. Our method outperforms other methods, even surpassing ModaLink, which operates on cropped point clouds. Fig. \ref{fig5:recall} shows the recall curves for different values of N, clearly illustrating our leading advantage.

\textbf{Robustness against random yaw angles.} When revisiting a location, the camera not only drifts in position but also has a random yaw angle. Thus, we randomly rotate the point clouds and report the results in \mbox{Table \ref{tab2:yaw}}. It can be seen that our method is rotation-invariant, thanks to the 360° multi-view descriptors for each point cloud, as illustrated in \mbox{Fig. \ref{fig6:rotation}}. In contrast, point cloud cropping can randomly omit some important scenes, thereby leading to a sharp drop in performance for ModaLink and I2P-Rec.

\textbf{Train on KITTI and test on KITTI-360.} We report the results in Table \ref{tab3:kitti360} to demonstrate the generalization of our method. Without fine-tuning, our method maintains high accuracy on KITTI-360 but falls short of the average reported in Table \ref{tab:recall-kitti}. Part of the decline is due to the longer distances of data sequences, which means retrieving from a larger database and thus increases the difficulty. To demonstrate this, we reduced the sequence length to $1/3$, making it closer to KITTI. It results in a significant improvement on performance compared to the original length.

\subsection{Ablation Studies}
\label{sec:ablation}
\textbf{Pixel-View-Scene Loss and xNetVLAD.} We conducted experiments by removing their individual components, including the three loss functions, Geometric Compensation (GEOC) and Efficient Inference (EFFI).  Table \ref{tab4:loss_netvlad} confirms the positive impact of each. Notably, all experiments rely on multi-view loss, as training cannot proceed without it.

\begin{table}[t]
\centering
\caption{Ablation studies on the number of views per point cloud.}
\vspace{-2mm}
\label{tab5:view_num}
\begin{tabular}{c|c|c}
\hline
           & KITTI-00                & KITTI-02                \\ \hline
$n_v$  & R@1\,/\,R@5\,/\,R@10\,/\,R@1\%      & R@1\,/\,R@5\,/\,R@10\,/\,R@1\%      \\ \hline
3  & 9.14\,\,/\,\,13.83\,\,/\,\,17.31\,\,/\,\,28.89 & 4.33\,\,/\,\,6.26\,\,/\,\,8.17\,\,/\,\,17.12 \\
5  & 79.81\,\,/\,\,90.24\,\,/\,\,95.00\,\,/\,\,98.99 & 55.93\,\,/\,\,68.85\,\,/\,\,77.97\,\,/\,\,92.62 \\
10 & 93.08\,\,/\,\,96.70\,\,/\,\,98.19\,\,/\,\,99.80 & 77.75\,\,/\,\,86.76\,\,/\,\,91.72\,\,/\,\,97.21 \\ 
15 & 93.24\,\,/\,\,97.00\,\,/\,\,98.19\,\,/\,\,99.78 & 77.71\,\,/\,\,86.46\,\,/\,\,91.72\,\,/\,\,97.30 \\ 
20 & 93.19\,\,/\,\,96.92\,\,/\,\,98.13\,\,/\,\,99.80 & 77.13\,\,/\,\,86.72\,\,/\,\,91.61\,\,/\,\,97.38 \\
30 & 93.13\,\,/\,\,96.83\,\,/\,\,98.15\,\,/\,\,99.74 & 76.98\,\,/\,\,86.44\,\,/\,\,91.72\,\,/\,\,97.51 \\ \hline
\end{tabular}
\end{table}

\begin{table}[t]
\centering
\caption{Ablation studies on the deep learning architecture.}
\vspace{-2mm}
\label{tab6:architecture}
\begin{tabular}{c|c|c|c|c|c}
\hline
Backbone &  T (ms)  &   M (GB)  &   P (M)  &   F (G)   &  R@1\,/\,R@1\% \\ \hline
   CNN   &  21.98  &   1.97   &   71.70 &  123.27  &  65.38\,\,/\,\,95.97   \\  
   ViT   &  410.9  &   19.92  &   41.74 &  133.99  &  N/A\,\,/\,\,N/A  \\  
   VMamba   &  25.31  &   1.79   &   50.79 &  112.58  &  93.13\,\,/\,\,99.74   \\  \hline
\end{tabular}
\footnotesize{The T, M, P and F represent Time, GPU Memory, Parameters and FLOPs, respectively. ViT encountered \textcolor{red}{out-of-memory} errors during training.\qquad}
\vspace{-3mm}
\end{table}

\textbf{The number of views per point cloud.} Table \ref{tab5:view_num} shows that fewer views lead to decreased place recognition accuracy due to potential loss of key scene information. Once the view density reaches a certain level, accuracy gains level off.

\textbf{The deep learning architectures.} To demonstrate the contribution of the visual state space model (VMamba) to our method, we implement the same dual-pyramid backbone (see \mbox{Fig. \ref{fig2}}) using CNN and ViT. With the same inputs, i.e., an RGB image of size $120\!\times\!600\!\times\!3$ and a 360° range image of size $48\!\times\!900\!\times\!1$, we report their running efficiency and performance in \mbox{Table \ref{tab6:architecture}}. Despite the memory overflow preventing ViT training, we still evaluated its efficiency, which falls far short of VMamba. Overall, the VMamba-based backbone performs as efficiently as CNNs, while greatly improving place recognition accuracy.

\section{Conclusion}
Towards monocular camera localization within pre-built LiDAR maps, this paper presents an efficient VMamba-based framework to generate global descriptors for RGB images and multi-view descriptors for point clouds. We propose a visible point based strategy to quantify the similarity between them, and a pixel-view-scene joint supervision for the cross-modal contrastive learning. The final multi-view matching significantly alleviates the sensor FOV differences. Additionally, the geometric compensation and efficient multi-view generation of xNetVLAD have also been proven effective in experiments. Future work will focus on exploring sequential representation and real-world applications.

\bibliographystyle{IEEEbib}
\bibliography{icra2025}

\end{document}